\documentclass[pdflatex,sn-basic,iicol]{sn-jnl}

\usepackage[caption=false]{subfig}
\usepackage{graphicx}%
\usepackage{multirow}%
\usepackage{amsmath,amssymb,amsfonts}%
\usepackage{amsthm}%
\usepackage{mathrsfs}%
\usepackage[title]{appendix}%
\usepackage{xcolor}%
\usepackage{textcomp}%
\usepackage{manyfoot}%
\usepackage{booktabs}%
\usepackage{algorithm}%
\usepackage{algorithmicx}%
\usepackage{algpseudocode}%
\usepackage{listings}%
\usepackage{silence}
\begin{document}

\title[Article Title]{DialogPaint: A Dialog-based Image Editing Model}



\author[1,2,3]{\fnm{Jingxuan} \sur{Wei}}\email{weijingxuan20@mails.ucas.edu.cn}
\equalcont{These authors contributed equally to this work.}

\author[1,2,4]{\fnm{Shiyu} \sur{Wu}}\email{wushiyu2022@ia.ac.cn}
\equalcont{These authors contributed equally to this work.}

\author[1]{\fnm{Xin} \sur{Jiang}}\email{jiangxin@baai.ac.cn}

\author*[1]{\fnm{Yequan} \sur{Wang}}\email{tshwangyequan@gmail.com}

\affil*[1]{\orgdiv{Beijing Academy of Artificial Intelligence}}

\affil[2]{\orgdiv{University of Chinese Academy of Sciences}}

\affil[3]{\orgdiv{Shenyang Institute of Computing Technology}, \orgname{Chinese Academy of Sciences}}

\affil[4]{\orgdiv{Institute of Automation}, \orgname{Chinese Academy of Sciences}}







\abstract{We introduce DialogPaint, a novel framework that bridges conversational interactions with image editing, enabling users to modify images through natural dialogue. By integrating a dialogue model with the Stable Diffusion image transformation technique, DialogPaint offers a more intuitive and interactive approach to image modifications. Our method stands out by effectively interpreting and executing both explicit and ambiguous instructions, handling tasks such as object replacement, style transfer, and color modification. Notably, DialogPaint supports iterative, multi-round editing, allowing users to refine image edits over successive interactions. Comprehensive evaluations highlight the robustness and versatility of our approach, marking a significant advancement in dialogue-driven image editing.}

\keywords{Dialogue-based Image Editing, Natural Language Processing, Image Transformation, Multi-round Interactions, Interactive Image Modifications}



\maketitle

\section{Introduction}\label{sec1}

Recently, great progress has been achieved in the field of image generation through the use of diffusion models \citep{RN3, RN4, RN5, RN8, RN9}. These large-scale text-to-image models have enabled the synthesis of high-quality, diverse images using concise textual prompts. Owing to their vivid output and stable training performance, diffusion models have surpassed generative adversarial networks (GAN) \citep{RN6} in popularity. Consequently, an increasing number of individuals are utilizing various diffusion models for a wide array of tasks, engaging in the creation of personalized images.

\begin{figure}
\begin{center}
\includegraphics[width=\linewidth]{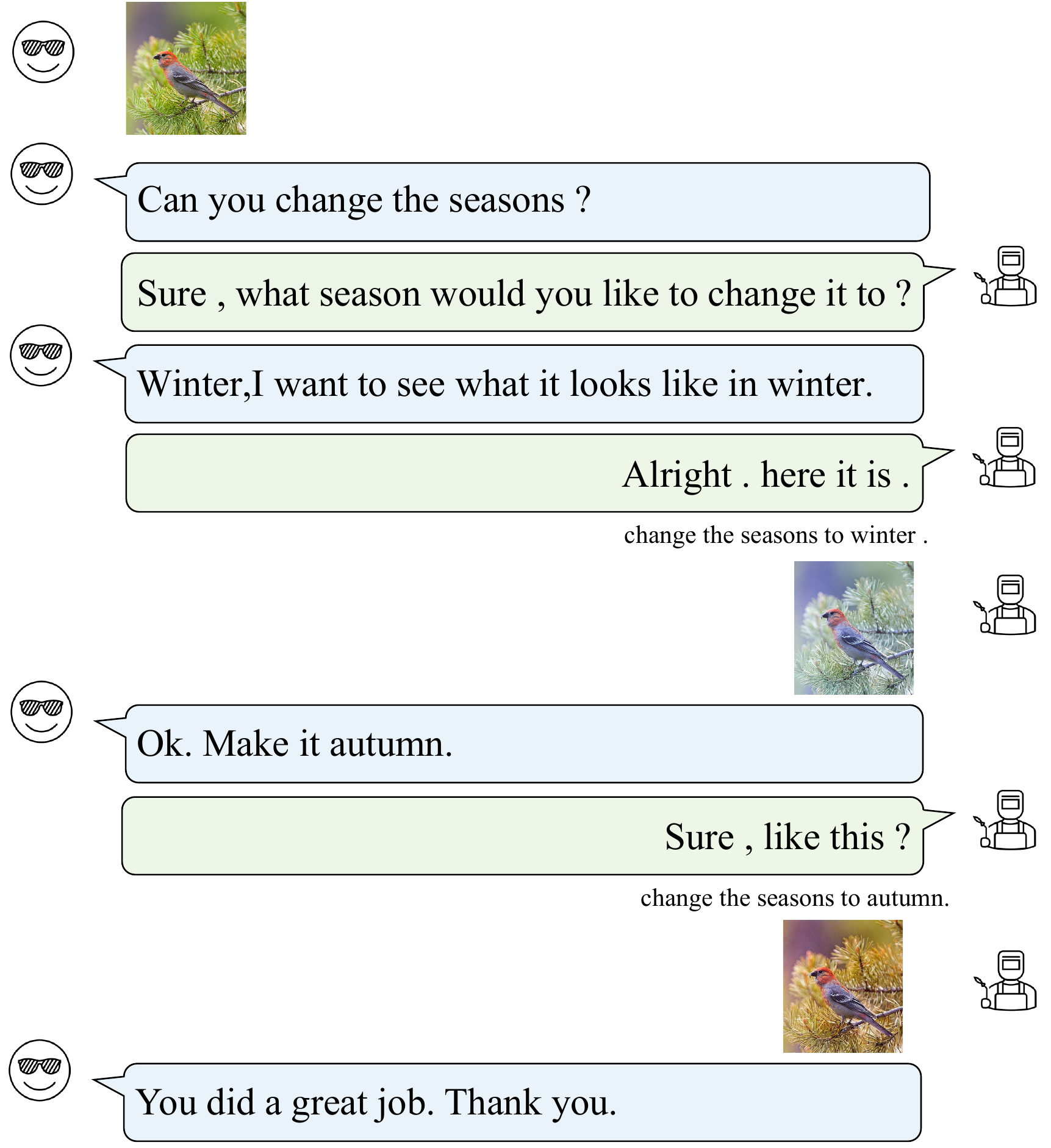}
\end{center}
\caption{An example of interactive editing}
\label{fig:img_bird}
\end{figure}

While advancements in image generation have led to results that are widely accepted and appreciated by the general public, other domains, such as image editing, still pose significant challenges. Our survey indicates that in the realm of image editing, users often lean towards refining an existing image rather than creating an entirely new one with similar content. This inclination is not merely a matter of convenience but is deeply rooted in human cognition. Modifying specific elements of an image resonates more naturally with our thought processes, reflecting a continuity of perception.

The allure of using natural language for guiding image edits is undeniable. It offers an intuitive interface, aligning with the way humans communicate. Despite the plethora of current methods that employ generative models for semantic image editing \citep{RN7}, many either lack a seamless human-computer interaction experience or demand extensive manual fine-tuning. Delving deeper into this issue, we discerned two primary bottlenecks:

\begin{enumerate}
    \item \textbf{Instruction Unfriendliness}: A significant portion of text-to-image models find it challenging to interpret and act upon human instructions. This limitation, which we term "instruction unfriendly," stems from their training predominantly on descriptive sentences. As a result, they are less adept at processing directive or instructional content.
    \item \textbf{Ambiguity in Instructions}: Users often provide models with ambiguous or vague directives. For instance, terms like "something else" without a clear reference can lead to confusion, especially when no reference image is provided. Existing models grapple with such imprecise instructions, leading to unsatisfactory edits.
\end{enumerate}

To mitigate these challenges, we advocate for a dialogue-driven approach. Engaging users in iterative interactions allows for refining and clarifying their instructions. Through this conversational paradigm, models can extract precise directives, ensuring more accurate image edits.

In this work, we present a holistic approach to image editing, harnessing the power of natural language dialogues. Users initiate the process by supplying an input image and a preliminary editing directive. Our integrated system, comprising a dialogue model \citep{RN11} and an image generation model \citep{RN9}, facilitates a conversation with the user. If the initial directive is ambiguous, the dialogue model poses clarifying questions, refining the instruction iteratively. Once a clear directive is established, the image generation model applies the desired edits. Recognizing the absence of suitable datasets for training such a system, we employ a self-instruct methodology \citep{RN29}, generating synthetic dialogues and image pairs for fine-tuning. Our evaluations validate the efficacy of our approach, showcasing its versatility in handling diverse editing tasks, from object replacement to style transfer. We also provide a publicly accessible demo, illustrating our system's interactive capabilities (refer to Figure \ref{fig:img_bird}).

Our contributions are summarized as follows:
\begin{itemize}
    \item We introduce an innovative framework that leverages interactive dialogue for image editing tasks.
    \item In pursuit of an enriched data foundation, we outline a novel approach to construct a dialogue-centric dataset specifically tailored for image editing.
    \item Through rigorous objective and subjective assessments, our framework exhibits robust capability in handling vague instructions and intricate image editing tasks.
\end{itemize}

\section{Related Work}\label{sec2}

\subsection{Large Language Models}
Large language models \citep{RN13, RN11, RN14, RN15, DBLP:journals/tmlr/WeiTBRZBYBZMCHVLDF22,DBLP:conf/icml/BidermanSABOHKP23,DBLP:journals/corr/abs-2302-13971} have been widely studied in recent years, with the capability to chat with humans fluently. Models such as GPT-3 \citep{RN15} can generate simulated data according to given samples, which is a convenient way to gather language data in a specific format and fine-tune other language models. Furthermore, conversation-oriented language models have also received attention.

DialoGPT \citep{DBLP:conf/acl/ZhangSGCBGGLD20} is an open-domain conversation model that generates high-quality human-like responses in conversations. Meena \citep{DBLP:journals/corr/abs-2001-09977} is a chatbot developed by Google that aims to train more conversations and empathy in human interactions. BlenderBot \citep{DBLP:conf/eacl/RollerDGJWLXOSB21} is a conversation agent trained on different conversation datasets from social media platforms, and it can converse on a wide range of topics. ChatGPT \citep{DBLP:journals/corr/abs-2303-08774} is a large-scale language model developed by OpenAI for generating high-quality text in a conversational context. It has shown exceptional performance on various conversational tasks. As the field continues to evolve, a plethora of new large language models are emerging, further pushing the boundaries of what's possible in natural language processing.

\subsection{Diffusion Models}

Diffusion model \cite{RN3, RN4, RN5, RN8, RN9} is a new kind of generative models which generate images from Gaussian noise by progressively denoising it. 
The model gradually adds noise to the input image by a preseted noise adding method which is named as forward process. And 
then it uses a deep neural network to restore the original image, which is called sampling process. As the dimensions of 
latent space in diffusion models can be really high, the output images can be very fantastic with high quality and diversity. 
Usually, a diffusion model is trained on the following variant of the variational bound:
\begin{equation}
    L_{simple} = \mathbb{E}_{\mathbf{I}_0, \boldsymbol{c}, \boldsymbol{\epsilon}, t} \left( || \boldsymbol{\epsilon} - 
    \boldsymbol{\epsilon}_\theta (\mathbf{I}_0, t, \boldsymbol{c}) ||^2_2 \right)
\end{equation}
where $\mathbf{I}_0$ and $\boldsymbol{c}$ are input images and optional conditions, $t \sim \mathcal{U} (0, 1)$ are time 
steps and $\boldsymbol{\epsilon} \in \mathcal{N} (0, 1)$ are added gaussian noise in forward process. $\boldsymbol{\epsilon}_
\theta$ is a learnable neural network which predicts the noise added on the image of previous moment. Usually a UNet \cite{RN16} is used 
to quickly and efficiently do this job. conditions $\boldsymbol{c}$ are the semantic embeddings of the input sentences processed 
by CLIP \cite{RN17}. With sampling methods such as DDIM \cite{RN5} and DPM-Solver \cite{RN18, RN19} that can speed up the sampling process Conspicuously, diffusion models 
is able to synthesize an image in $15 \sim 25$ steps. The widly used stable diffusion applies an AutoEncoder, which encodes the input image $\mathbf{I}$ into a latent space first and decode the sampling output to an real image. In this case, diffusion models 
will not deal with high-frequency details and can synthesis images with higher quality. 

\subsection{Text-driven Image Editing}

Text-driven editing with GAN \citep{RN20, RN21, RN22, RN31, RN32} has been carefully studied in recent years. Early works targeted at single task like style transferring. 
They trained the model with special image pairs to complete special editing tasks, which is based on domain transferring. With the 
advent of CLIP, now people can guide image editing with texts as input conditions. As for diffusion models, some of them \citep{RN23, RN30} natively 
have the ability for editing images due to the strong capabilities of text features extraction by CLIP\citep{RN17}. Models with mask guidance \citep{RN28} can 
make the edit more accurately. Another editing way is using textual inversion \citep{RN25, RN24}. Models will learn a speical word in textual embedding 
space and bind it with the specific subject in the given image. After training, with a sentence that contains the special word, the 
diffusion model can generate images of the specific subject in different scenes described by the sentence.

\section{Methodology}\label{sec3}
\begin{figure*}[htbp]
\begin{center}
\includegraphics[width=\linewidth]{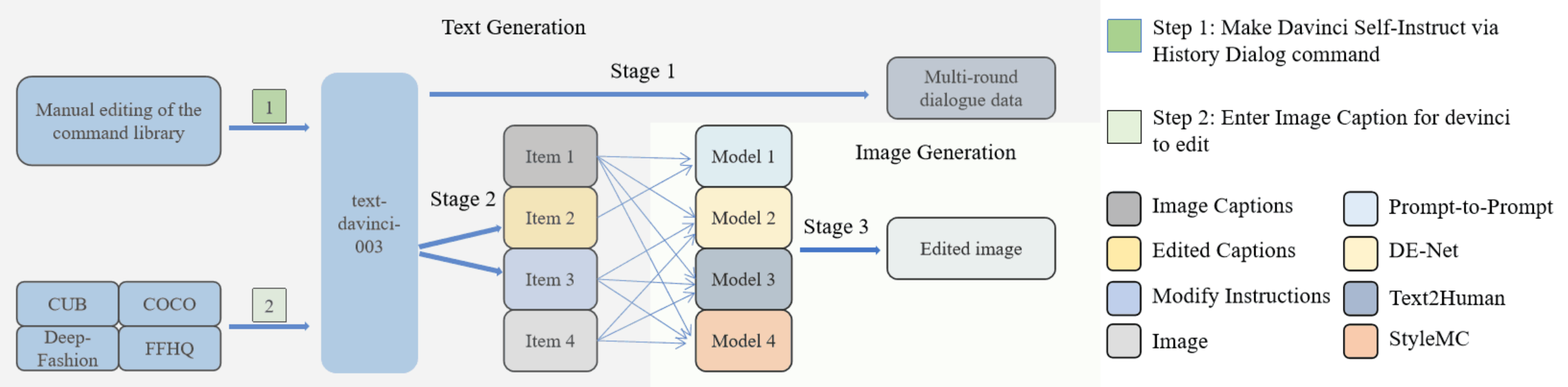}
\end{center}
\caption{The processing of dialogue and image editing dataset construction}
\label{fig:all_ataset}
\end{figure*}
DialogPaint accepts user's images and editing instructions in the form of dialogue. It implements instruction clarification for vague instructions, and provides feedback and summaries to complete dialogue-based image editing. The core of this task lies in the construction of a dialogue-based image editing dataset (Sec. \ref{3.1}), and the construction of a dialogue-based image editing model (Sec.\ref{3.2}).

\subsection{Construction of Dialogue and Image Editing Datasets}
\label{3.1}

\subsubsection{Building Dialogue Dataset}
\label{3.1.1}

\begin{figure}[t]
\begin{center}
\includegraphics[width=\linewidth]{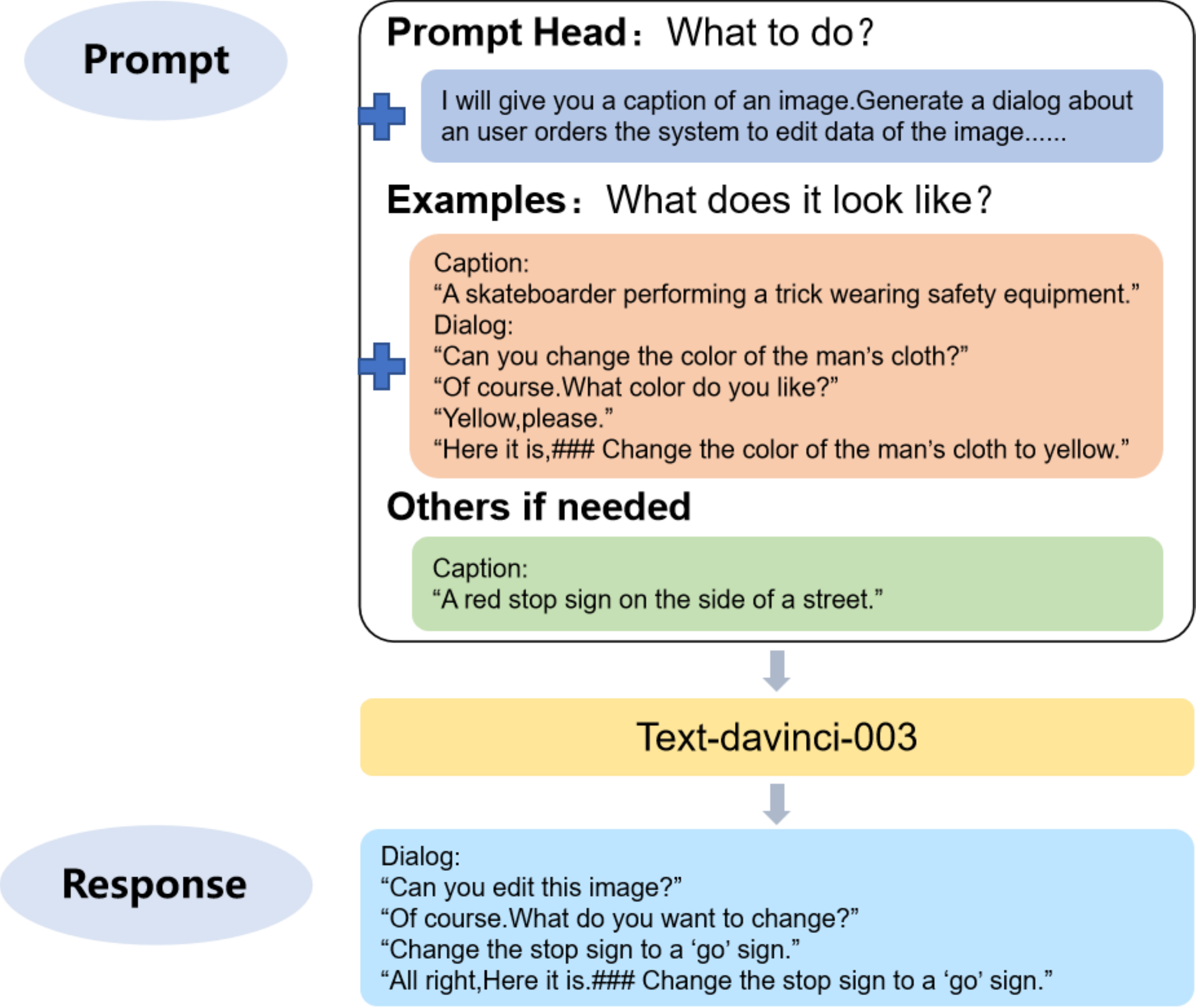}
\end{center}
\caption{Example of Dialogue Dataset Construction }
\label{fig:Dialogue_Dataset}
\end{figure}
The construction of both the dialogue and image editing datasets is depicted in Figure \ref{fig:all_ataset}. This process encompasses two primary steps: building the dialogue dataset and the image editing dataset. For the dialogue dataset, image captions were randomly sourced from four distinct datasets: CUB-200-2011 \citep{DBLP:journals/tcsv/HeP20}, Microsoft COCO \citep{DBLP:conf/eccv/LinMBHPRDZ14}, DeepFashion \citep{DBLP:conf/cvpr/LiuLQWT16}, and Flickr-Faces-HQ (FFHQ) \citep{DBLP:journals/pami/KarrasLA21}. These captions were then merged with prompt instructions to form the necessary dialogue data. For the image editing dataset, image-text pairs were randomly chosen from the aforementioned datasets and processed using text-davinci-003 to generate editing instructions. The selection of these datasets was motivated by their extensive diversity and prominence in the computer vision domain.

For the dialogue dataset, 10,000 image captions were randomly selected from the four aforementioned datasets. Using the self-instruct tool \citep{RN29}, these image captions were amalgamated with prompt instructions and inputted into text-davinci-003 to produce the required dialogue data, as illustrated in Step 1 of Figure \ref{fig:all_ataset}.

Figure \ref{fig:Dialogue_Dataset} provides an example of the dialogue generation task defined using a Prompt Head. The task involved instructing text-davinci-003 to "generate a dialogue about a user instructing the system to edit the image based on the given image caption." Subsequently, 20 manually crafted dialogue examples were randomly selected from a library containing 500 dialogue examples, as shown in the Example section of Figure \ref{fig:Dialogue_Dataset}. This section consists of two parts: the "Caption," representing the input image caption, and the "Dialog," which depicts the desired dialogue. This desired dialogue simulates multiple rounds of conversation aimed at altering the image's attributes, such as colors and scenes, and clarifying ambiguous instructions. Ultimately, an image caption was added to the Examples section and input into text-davinci-003. The resulting response from text-davinci-003 was used to craft the desired dialogue data.

Using the aforementioned method, we constructed a dialogue dataset comprising 10,000 dialogue data samples. Manual inspection of these samples confirmed their appropriateness for open-domain dialogue image editing tasks.

\subsubsection{Building Image Editing Dataset}

The creation of the image editing dataset, detailed in Step 2 and Stage 2 of Figure \ref{fig:all_ataset}, involves two stages. Initially, image editing instructions are generated (Step 2), followed by the production of edited images using text-to-image editing models based on the instructions from Step 2 (Stage 2). We randomly selected 10,000 image-text pairs from the CUB-200-2011, Microsoft COCO, DeepFashion, and FFHQ datasets. These pairs were processed by text-davinci-003, using the self-instruct tool \citep{RN29} to combine image captions and prompt instructions, resulting in the desired text editing instructions. The generated text data was then input into text-to-image editing models in Stage 2 to produce the edited images.

\begin{figure}[t]
\begin{center}
\includegraphics[width=\linewidth]{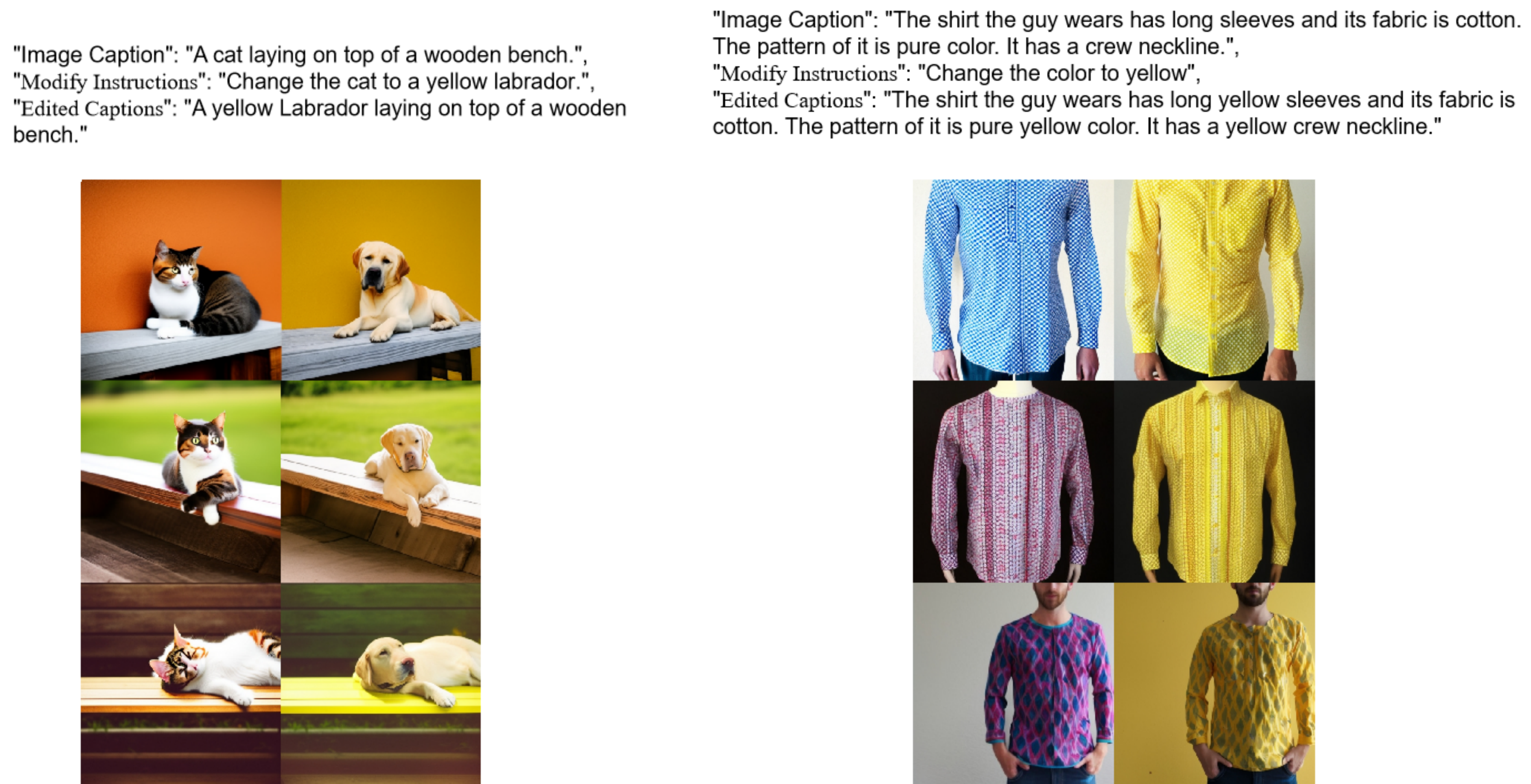}
\end{center}
\caption{Example of Image Editing Dataset}
\label{fig:Image_Editing_Dataset}
\end{figure}

During Step 2, similar to Figure \ref{fig:Dialogue_Dataset}, we defined a Prompt Head to articulate the task. Text-davinci-003 was instructed to generate modification instructions based on the caption. We selected 20 human-crafted editing instructions from a sample library of 500 instances and used them as depicted in Figure \ref{fig:Image_Editing_Dataset}. We input an image caption along with the selected examples into text-davinci-003, yielding the desired Modify Instructions and Edited Captions data.

Datasets produced in this phase were utilized for fine-tuning the model for image object isolation and partial transformation, as detailed in Section \ref{3.2.2}.

During Stage 2, the editing instructions from Step 2 were combined with pre-trained models to produce edited images. We employed four text-to-image editing models, including Prompt-to-Prompt \citep{RN26}, DE-Net \citep{DBLP:journals/corr/abs-2206-01160}, Text2Human \citep{DBLP:journals/tog/JiangYQWLL22}, and StyleMC \citep{DBLP:conf/wacv/KocasariDTY22}, following prior works. As depicted in Figure \ref{fig:all_ataset} Stage 2, Prompt-to-Prompt uses Image Captions and Edited Captions to generate original and edited images, while DE-Net, Text2Human, and StyleMC employ Image Captions, Modify Instructions, and the original image for the same purpose. After generation, images were filtered using a CLIP-based metric to assess the similarity change between original and edited images, ensuring dataset quality.

Finally, after manual review, we obtained 6468 pairs of original and edited image-text pairs that meet the criteria for open-domain image editing tasks.

\label{3.1.2}

\subsection{Construction of Dialogue and Image Editing Models}
\label{3.2}
\begin{figure*}[htbp]
\begin{center}
\includegraphics[width=\linewidth]{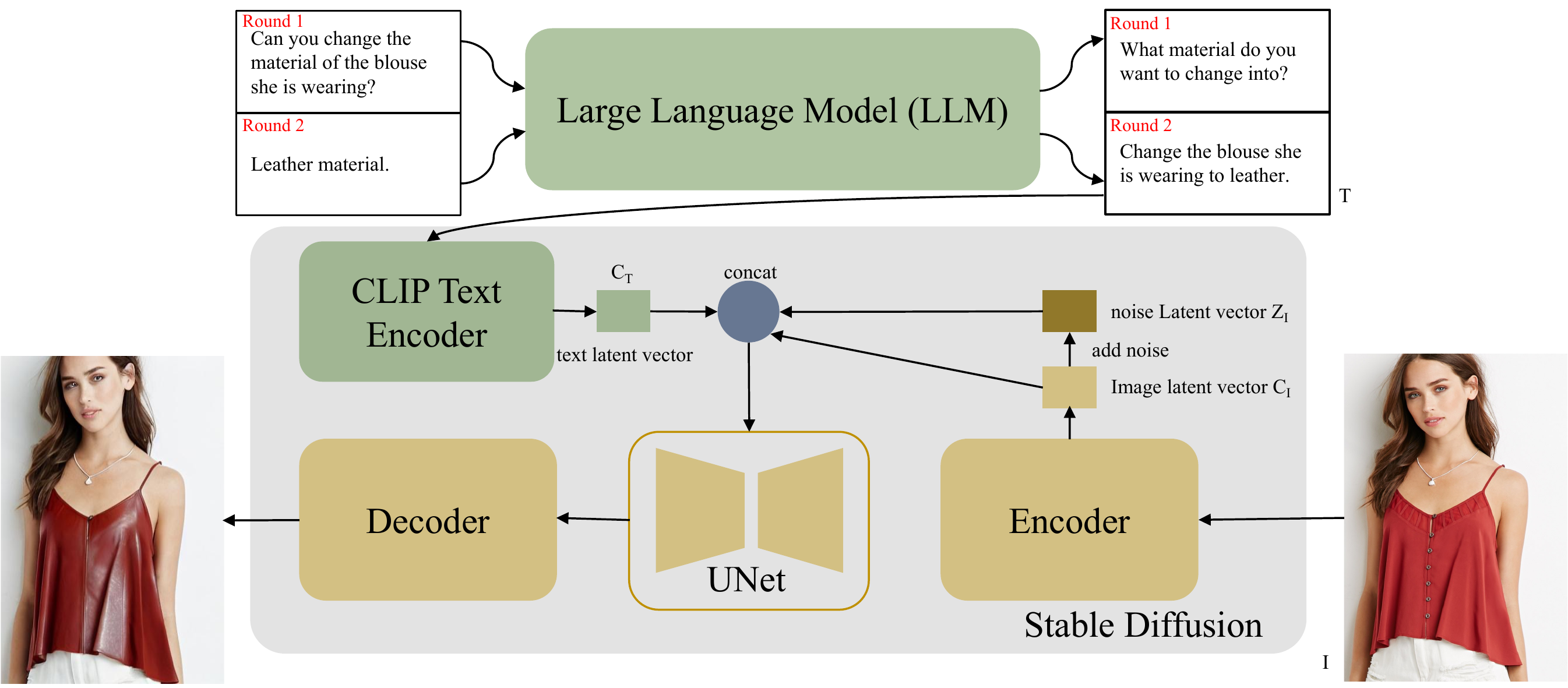}
\end{center}
\caption{Model Architecture for Dialogue-Based Image Editing}
\label{fig:img_model}
\end{figure*}

DialogPaint is architected as a cohesive system, integrating dialogue-driven interactions with sophisticated image editing capabilities. The system is bifurcated into two primary components: the Dialogue Model and the Image Editing Model. The overarching design philosophy is to first harness the Dialogue Model to generate context-aware dialogue responses tailored for image editing tasks. Subsequently, the Image Editing Model is invoked to perform image editing based on the explicit textual instructions derived from the dialogue interactions. The overall architecture of the system is illustrated in Figure \ref{fig:img_model}.

\subsubsection{Dialogue Model Construction}
\label{3.2.1}
The Dialogue Model is pivotal to our approach, designed to facilitate open-domain conversations for image editing. Given a natural language prompt describing an image and an editing task, the model is trained to generate a series of dialogue responses. The culmination of these interactions is a clear and precise instruction, denoted as \( T \), for the Image Editing Model.

To achieve this, we utilized the pre-trained weights from the Blender dialogue model \citep{DBLP:conf/eacl/RollerDGJWLXOSB21}. The primary objective during this phase was to adapt the model to generate high-quality, contextually relevant responses that can provide explicit editing instructions. This adaptation ensures that the model is attuned to the nuances of image editing dialogues, making it capable of understanding intricate user requirements and generating appropriate responses. The model's capability to generate such precise instructions is crucial, as it bridges the gap between user intent and the subsequent image editing operations performed by the Image Editing Model.

\subsubsection{Image Editing Model Construction}
\label{3.2.2}
The Image Editing Model is devised to interpret the explicit textual instructions \( T \) from the Dialogue Model and enact the corresponding image edits on the input image \( I \). The textual instruction \( T \) is encoded through CLIP to obtain a latent vector representation \( C_T \). Concurrently, the input image \( I \) is processed through an encoder to derive its latent representation \( C_I \). Noise is subsequently introduced to \( C_I \) to produce \( Z_I \), which is then subjected to the diffusion process.

Our approach employs the Stable Diffusion architecture, leveraging the foundational principles from InstructPix2Pix, to execute image editing based on the explicit textual instructions. The diffusion process is articulated as:

\begin{equation}
\resizebox{0.95\linewidth}{!}{$
L = \mathbb{E}_{z \sim q(z), C_I, C_T, \epsilon \sim \mathcal{N}(0,1), t} \left[ \| \epsilon - \epsilon_\theta(Z_I, t, C_I, C_T) \|_2^2 \right]
$}
\end{equation}

where \( q(z) \) denotes the posterior distribution of the image latents.

Furthermore, to enhance the model's adaptability to diverse editing instructions, we introduced an unconditional diffusion guidance mechanism. This mechanism employs two guidance scales, \(s_I\) and \(s_T\), to balance the alignment between the input image and the editing instruction. The modified score estimation is articulated as:
\begin{equation}
\resizebox{0.95\linewidth}{!}{$
\begin{aligned}
\tilde{e_\theta}\left(Z_I, C_I, C_T\right) &= e_\theta\left(Z_I, \varnothing, \varnothing\right) \\
&+ s_I \cdot \left(e_\theta\left(Z_I, C_I, \varnothing\right) - e_\theta\left(Z_I, \varnothing, \varnothing\right)\right) \\
&+ s_T \cdot \left(e_\theta\left(Z_I, C_I, C_T\right) - e_\theta\left(Z_I, C_I, \varnothing\right)\right)
\end{aligned}
$}
\end{equation}

\section{Experiments}\label{sec_4}
\subsection{Experimental Setup}
Our experimental framework was deployed on two distinct datasets: a dialogue dataset comprising 10,000 samples and an image editing dataset with 6,468 meticulously curated samples.

For the dialogue modeling phase, the dataset was partitioned into 9,000 training samples, with 500 samples each designated for validation and testing. We initialized our dialogue model using the pre-trained weights from the Blender dialogue model \citep{DBLP:conf/eacl/RollerDGJWLXOSB21}. The model was trained on 8 Nvidia Tesla A100 40G GPUs. The training hyperparameters included a batch size of 128, embedding size of 2560, ffn size of 10240, n-heads at 32, n-positions at 128, n-encoder-layers at 2, n-decoder-layers at 24, and dropout at 0.1. The Adam optimizer was used with a learning rate of 7e-06, and an early stopping mechanism was implemented, which halted training if no performance improvement was observed over 10 consecutive iterations.

For the image editing component, we initialized our model using the weights from InstructPix2Pix \citep{RN26}. The model was adapted to our image editing dataset using 8 Nvidia Tesla A100 40G GPUs. The hyperparameters included a batch size of 32 and an input image size of 256. Necessary adjustments were made to align with our dialogue-driven approach, and the model was trained over 125 epochs.

Our model architecture primarily consists of a large language model-guided dialogue module and an image editing module, as depicted in Figure \ref{fig:img_model}.

\subsection{Qualitative Analysis of Experimental Cases}

\begin{figure}[t]
\begin{center}
\includegraphics[width=\linewidth]{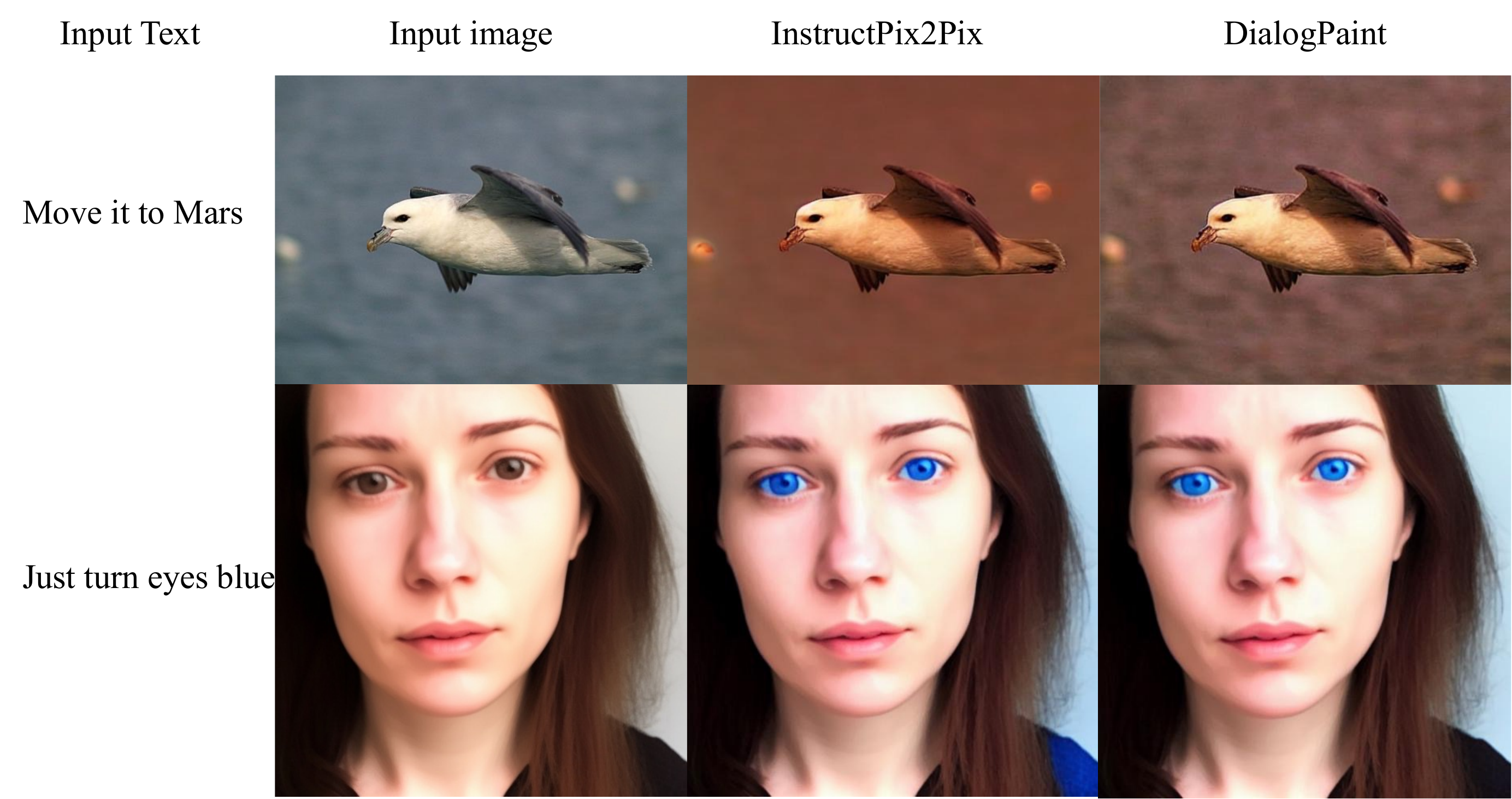}
\end{center}
\caption{Performance Contrast: DialogPaint vs. InstructPix2Pix with Identical Single-Turn Instructions}
\label{fig:img_comp}
\end{figure}

\begin{figure*}[htbp]
\begin{center}
\includegraphics[width=\linewidth]{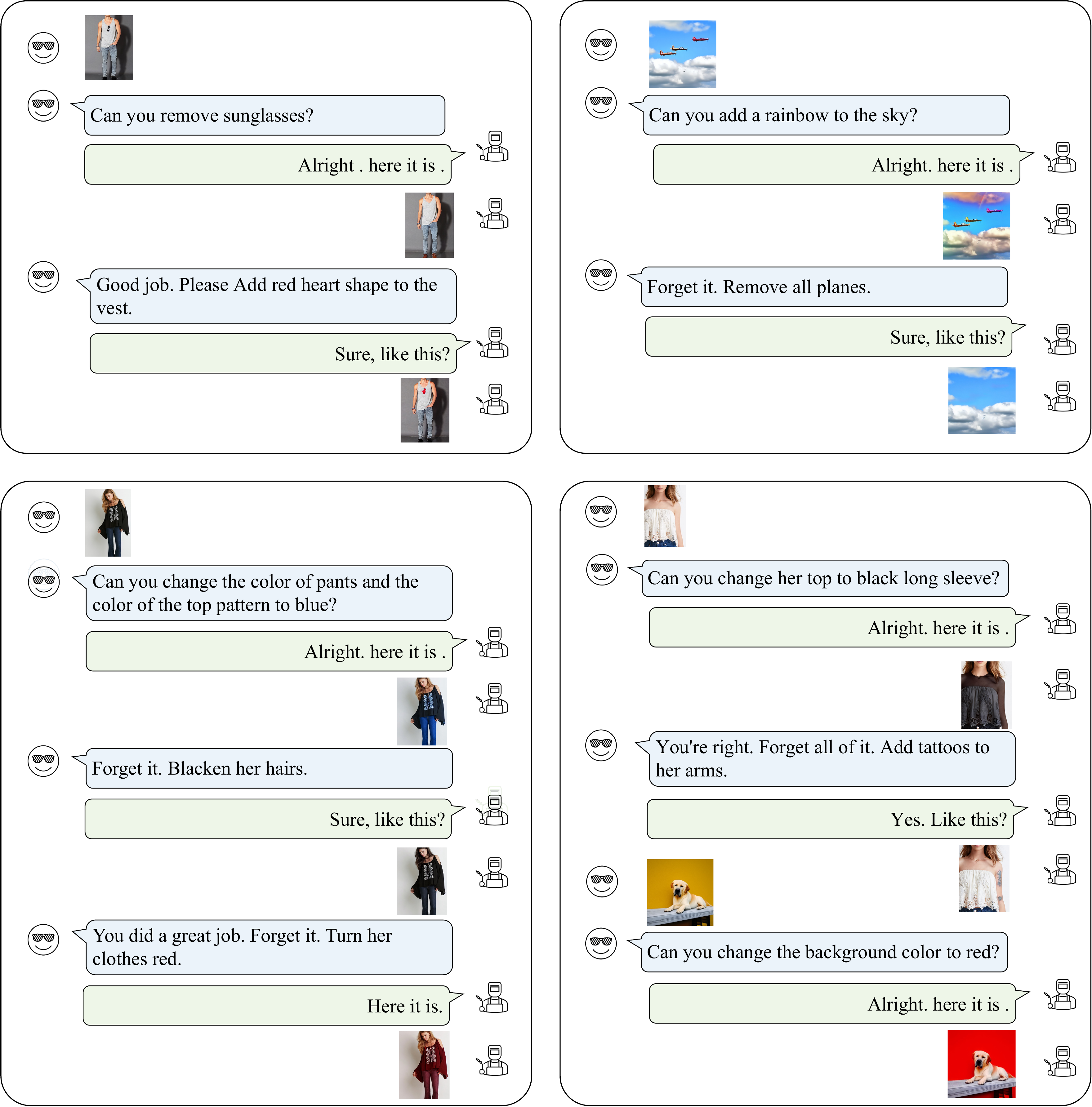}
\end{center}
\caption{Comprehensive Demonstration of DialogPaint's Multi-Turn Editing Capabilities}
\label{fig:img_examples}
\end{figure*}

Figure \ref{fig:img_comp} showcases the superior performance of our DialogPaint model in comparison to the baseline model, InstructPix2Pix, when provided with identical single-turn instructions. Notably, InstructPix2Pix exhibits overfitting tendencies, especially during substantial image transformations. This leads to a loss of intricate background details. In contrast, DialogPaint adeptly retains detailed background information, ensuring a more realistic transformation. The underlying architecture and training strategy of DialogPaint, which emphasizes context preservation and object-specific edits, contribute to this enhanced performance. For instance, with the instruction "Move it to Mars", DialogPaint's rendition appears more authentic, preserving the original image's ocean contours while adapting the background to resemble Martian terrain. When given the directive "Just turn eyes blue", DialogPaint demonstrates superior object isolation capabilities, avoiding unintended alterations.

Figure \ref{fig:img_examples} offers a holistic view of DialogPaint's capabilities across various editing scenarios. The top-left quadrant illustrates the model's proficiency in multi-turn dialogues, with successive edits like removing glasses and adding a red heart shape to the vest. The top-right quadrant emphasizes the model's scene transformation skills, with harmonious integration of new elements like rainbows. The "Forget" operation, which effectively undoes prior edits, is also demonstrated, underscoring the model's adaptability. The bottom-left quadrant highlights the model's finesse in color modifications, with realistic shadow and lighting adjustments. Lastly, the bottom-right quadrant displays the model's prowess in fine-grained edits and background color transitions, effectively capturing subtle changes in human skin tones and animal fur in response to surrounding color variations.

Our model's robustness is attributed to our meticulously curated dataset and the innovative integration of dialogue-driven context into the image editing process. The dataset's detailed images, paired with clear descriptions, empower the model to assimilate intricate knowledge and execute precise transformations. However, it's worth noting that while DialogPaint excels in many scenarios, there might be specific edge cases or complex multi-turn dialogues where further refinements could be beneficial. Through these demonstrations, DialogPaint's potential in delivering precise, dialogue-driven image edits across diverse scenarios is evident.

\subsection{Quantitative Analysis of Evaluation Metrics}
\begin{table}[t]
  \centering
  \caption{Evaluation Metrics for the Dialogue-based Image Editing Model}
    \begin{tabular}{lc}
    \toprule
    Evaluation Metric & Score \\
    \midrule
    Perplexity (ppl) & 1.578 \\
    Fréchet Inception Distance (FID) & 1.52 \\
    Precision-Recall Distance (PRD) & 1.56 \\
    Overall Satisfaction & 4.22 \\
    Mean Opinion Score (MOS) & 4.32 \\
    \bottomrule
    \end{tabular}%
  \label{tab:addlabel}%
\end{table}%

To rigorously assess the performance of our dialogue-based image editing model, we employed a combination of objective and subjective metrics, the results of which are tabulated in Table \ref{tab:addlabel}.

Objective Metrics:
- **Perplexity (ppl)**: This metric evaluates the model's proficiency in predicting subsequent words in a dialogue sequence. A lower perplexity score indicates a model's better understanding and prediction capability in dialogues.
- **Fréchet Inception Distance (FID)**: FID measures the similarity between generated images and real images in terms of their statistics. A lower FID score signifies that the generated images are of higher quality and more similar to real images.
- **Precision-Recall Distance (PRD)**: PRD evaluates the likeness between the distributions of generated and real images in a feature space. A lower PRD score indicates a better alignment between the distributions, suggesting more realistic image generation.

Subjective Evaluation:
We sought feedback from 100 participants who interacted with our dialogue editing model. Their overall experience was gauged on a scale of 1 to 5. The average rating for overall satisfaction was 4.22, indicating a high degree of user satisfaction with the model's performance. Additionally, the quality of the generated images was rated on the Mean Opinion Score (MOS) scale, with the average score being 4.32, further attesting to the model's capability in producing high-quality images.

In conclusion, our dialogue-driven image editing model showcases robust performance across both objective and subjective evaluation metrics, highlighting its potential for diverse applications.

\subsection{Model Limitations}\label{sec5}
\begin{figure}[t]
\begin{center}
\includegraphics[width=\linewidth]{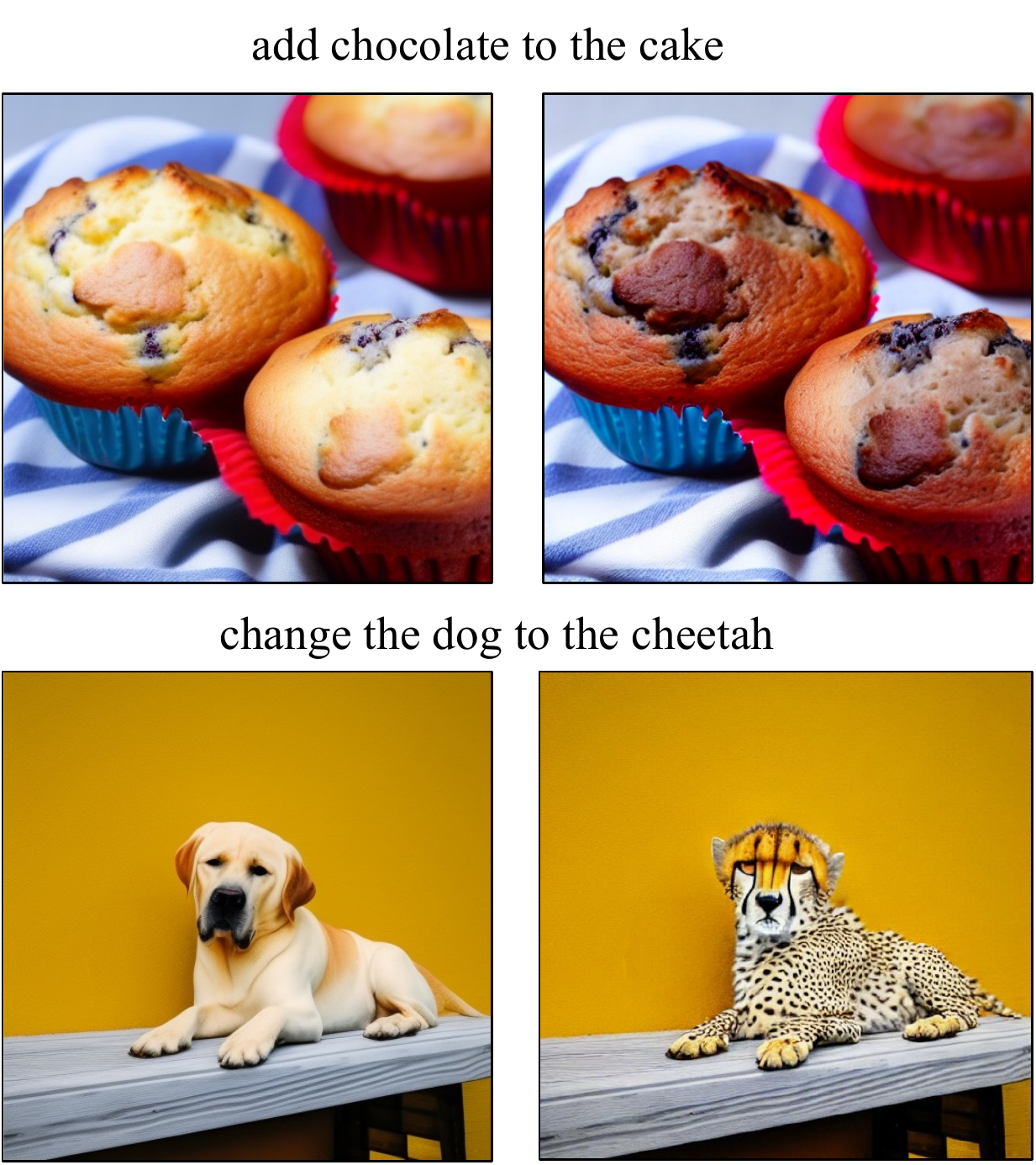}
\end{center}
\caption{Challenging Scenarios for DialogPaint: Illustrative Cases of Imperfect Edits}
\label{fig:fail_img}
\end{figure}

While our dialog-based image editing model, DialogPaint, has demonstrated promising capabilities in stylistic and color modifications, it is not without its limitations. These constraints primarily arise from the limited diversity and volume of dialog samples and image editing operations in our dataset.

In certain scenarios, especially when dealing with intricate image content, the model's edits may not fully align with user expectations. For instance, in the example depicted in Figure \ref{fig:fail_img} (top), the instruction "add chocolate to the cake" resulted in the addition of chocolate. However, the placement was not aesthetically pleasing, and the chocolate overlay obscured some elements of the original cake, such as the small cookies. In another example (bottom), the command "change the dog to the cheetah" led to a transformation that, while reminiscent of a cheetah, still retained some canine features. This highlights the model's current challenge in striking a balance between transformation and preservation, a nuance that is crucial for more natural and intuitive image edits.

Future endeavors will be directed towards refining the model's ability to interpret and act upon dialog-based editing instructions more accurately, especially in scenarios demanding a delicate balance between transformation and preservation.

\section{Conclusion}\label{sec6}

In this paper, we introduced DialogPaint, a novel dialogue-based image editing model that stands out by enabling image modifications through explicit instructions within a conversational context. Unlike traditional methods, our approach seamlessly integrates the power of natural language processing with image editing, providing a more intuitive and interactive user experience.

We constructed a unique dataset containing both dialogue and image editing samples, which played a pivotal role in training our model to understand and execute user instructions effectively. Our experimental results, both qualitative and quantitative, underscore the model's capability to perform image editing tasks across various domains, demonstrating its superiority over conventional methods.

However, as with any pioneering approach, our model has its limitations, especially when handling intricate editing tasks due to the constrained volume of dialog samples and image editing operations in our dataset. We have transparently discussed these limitations and provided visual examples for clarity.

Looking ahead, we envision refining our dataset to include more diverse dialog samples and operations, enhancing the model's robustness and versatility. Furthermore, we aim to explore the broader applications of dialogue-based image editing models in areas such as smart homes and facial recognition, pushing the boundaries of what's possible in the intersection of natural language processing and image editing.

In conclusion, DialogPaint represents a significant step forward in the realm of image editing, offering a fresh perspective on how we perceive and interact with image editing tasks. We believe our work lays a solid foundation for future research in this exciting interdisciplinary domain.

\section*{Acknowledgments}
This work is supported by the National Key R\&D Program of China (2022ZD0116300) and the National Science Foundation of China (NSFC No. 62106249).

\section*{Declarations}

\begin{itemize}
    \item \textbf{Funding:} This work is supported by the National Key R\&D Program of China (2022ZD0116300) and the National Science Foundation of China (NSFC No. 62106249).
    
    \item \textbf{Conflict of Interest/Competing Interests:} The authors declare that they have no competing interests.
    
    \item \textbf{Ethics Approval:} Not applicable.
    
    \item \textbf{Consent to Participate:} Not applicable.
    
    \item \textbf{Consent for Publication:} Not applicable.
    
    \item \textbf{Availability of Data and Materials:} We plan to make our dataset publicly available at an appropriate time in the future.
    
    \item \textbf{Code Availability:} The code used in this study will be made available upon reasonable request.
    
    \item \textbf{Authors' Contributions:} All authors contributed equally to the research, writing, and revision of this manuscript.
\end{itemize}


\bibliography{sn-bibliography}

\end{document}